\documentclass[letterpaper]{article} 
\usepackage{aaai24}  
\usepackage{times}  
\usepackage{helvet}  
\usepackage{courier}  
\usepackage[hyphens]{url}  
\usepackage{graphicx} 
\usepackage{natbib}  
\usepackage{caption} 
\usepackage{algorithm}
\usepackage{url}         
\usepackage{booktabs}      
\usepackage{amsfonts}    
\usepackage{nicefrac}     
\usepackage{microtype}   
\usepackage{xcolor}   
\usepackage{adjustbox}
\usepackage{graphicx}

\usepackage{bm}
\usepackage{epstopdf}
\usepackage{amsfonts}
\usepackage{multirow}
\usepackage{color}
\usepackage{fontenc} 
\usepackage{subcaption}
\usepackage{lineno}
\usepackage{booktabs}
\usepackage{comment}
\usepackage{mathtools}
\usepackage{algorithm,algpseudocode}
\usepackage{amsthm}
\usepackage{mathtools}
\usepackage{amsfonts,amssymb}
\usepackage{enumitem}

\newcommand{\model}{ {L2R-GNN} }
\newcommand{\framework}{ {Learning to Reweight for Generalizable Graph Neural Network }}

\newcommand{\indep}{\perp \!\!\! \perp}
\theoremstyle{definition}

\theoremstyle{remark}

\theoremstyle{proposition}

\usepackage{newfloat}
\usepackage{listings}
\DeclareCaptionStyle{ruled}{labelfont=normalfont,labelsep=colon,strut=off} 
\floatstyle{ruled}
\newfloat{listing}{tb}{lst}{}
\floatname{listing}{Listing}
\title{Learning to Reweight for Graph Neural Network}
\author{
Zhengyu Chen\textsuperscript{\rm 1,2}, Teng Xiao\textsuperscript{\rm 3}, Kun Kuang\textsuperscript{\rm 1,2}\thanks{Corresponding author}, Zheqi Lv\textsuperscript{\rm 1,2}, Min Zhang\textsuperscript{\rm 1,2}, Jinluan Yang\textsuperscript{\rm 1,2},  \newline Chengqiang Lu\textsuperscript{\rm 4}, Hongxia Yang\textsuperscript{\rm 4}, and Fei Wu\textsuperscript{\rm 1,2}
}
 
\affiliations{
    \textsuperscript{\rm 1} Institute of Artificial Intelligence, Zhejiang University\\
    \textsuperscript{\rm 2} Shanghai Institute for Advanced Study, Zhejiang University\\
    \textsuperscript{\rm 3} The Pennsylvania State University\\
    \textsuperscript{\rm 4} DAMA Academy, Alibaba Group.\\
    $\{$kunkuang, wufei$\}$@zju.edu.cn, yang.yhx@alibaba-inc.com
}

\usepackage{bibentry}

\begin{document}

\maketitle

\begin{abstract}
Graph Neural Networks (GNNs) show promising results for graph tasks. However, existing GNNs' generalization ability will degrade when there exist distribution shifts between testing and training graph data.
	The cardinal impetus underlying the severe degeneration is that the GNNs are architected predicated upon the I.I.D assumptions. In such a setting,  GNNs are inclined to leverage imperceptible statistical correlations subsisting in the training set to predict, albeit it is a spurious correlation.  
	In this paper, we study the problem of the generalization ability of GNNs in Out-Of-Distribution (OOD) settings.
	To solve this problem, we propose the \framework (L2R-GNN) to enhance the generalization ability for achieving satisfactory performance on unseen testing graphs that have different distributions with training graphs.
	We propose a novel nonlinear graph decorrelation method, which can substantially improve the out-of-distribution generalization ability and compares favorably to previous methods in restraining the over-reduced sample size.
 The variables of the graph representation are clustered based on the stability of the correlation, and the graph decorrelation method learns weights to remove correlations between the variables of different clusters rather than any two variables. 
	Besides, we interpose an efficacious stochastic algorithm upon bi-level optimization for the \model  framework, which facilitates simultaneously learning the optimal weights and GNN parameters, and avoids the overfitting problem. 
	Experimental results show that \model greatly outperforms baselines on various graph prediction benchmarks under distribution shifts.
\end{abstract}

\section{Introduction}
\label{sec:introduction}

Graph Neural Networks (GNNs) have achieved state-of-the-art performances on various graph tasks ~\cite{kipf2016semi, velivckovic2017graph, xu2018powerful}, but they assume that the training and testing data are independent and identically distributed (i.e., i.i.d assumption), which is not always the case in real-world applications \cite{chen2021deep,chen2021multi,chen2019deep,chen2021improving,gai2019deep}. This leads to inadequate out-of-distribution ({OOD}) generalization ability, causing significant performance degradation under distribution shifts~\cite{hu2020open, wu2018moleculenet,chen2021pareto}. 

\begin{figure*} 
	\centering
	\includegraphics[width=0.7\linewidth]{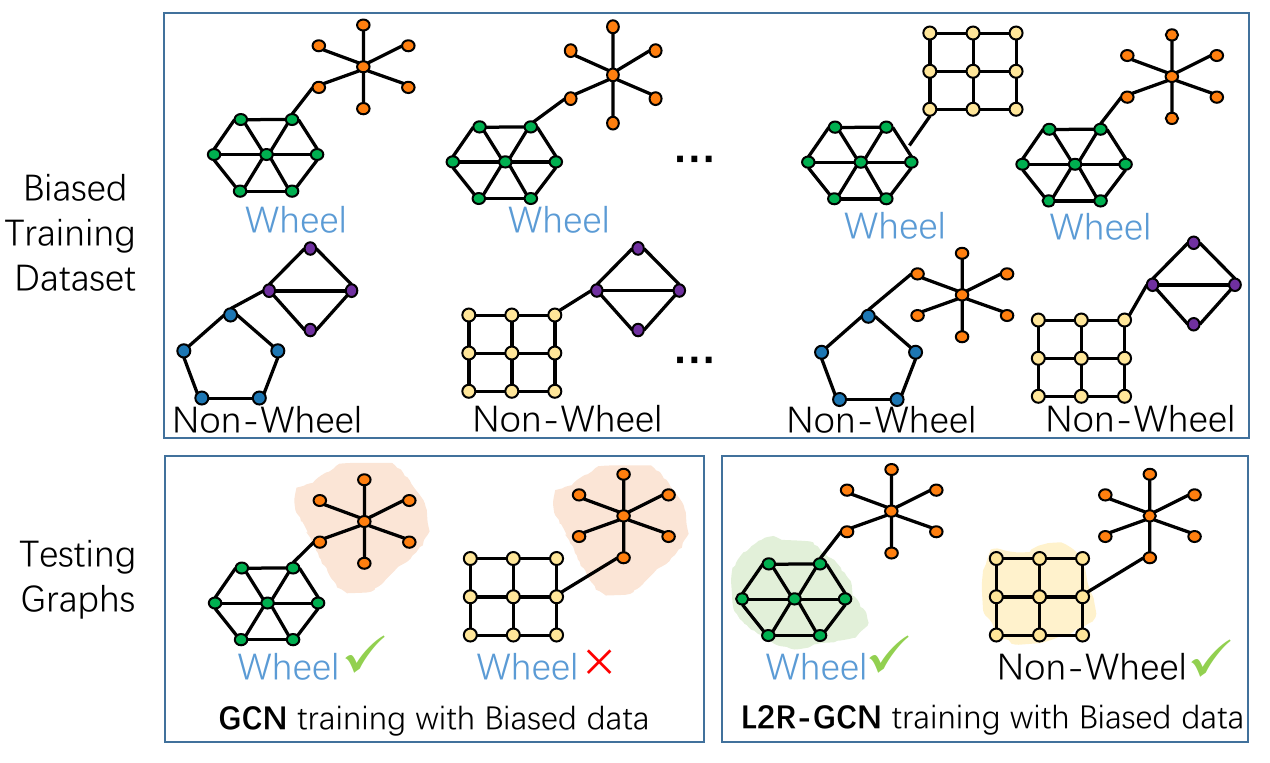}
	\vspace{-1em}
	\caption{An illustration of a fictitious correlation in the ``wheel" motif graph classification task.}
	\label{fig:fig1}
\end{figure*}

GNNs' inadequate out-of-distribution generalization is caused by a spurious correlation between irrelevant features and category labels in training data \cite{xiao2022representation,chen2023invariant}. This correlation varies across distributions and is exploited by GNNs for inference. An example of spurious correlation is shown in the graph classification task of the ``wheel" motif in Figure~\ref{fig:fig1}. In the biased training dataset, most positive graphs have only ``star" motifs added, leading to a strong correlation between structural features of ``wheel" motifs and ``star" motifs. This unexpected correlation leads to a spurious correlation between the structural features of ``star" motifs and the label ``wheel". The GCN model exploits this spurious correlation and tends to use ``star" motifs for prediction, making false predictions on negative graphs with a ``star" motif.

To solve the spurious correlation problem caused by the discrepancy between training and testing distributions, 
earlier research attempts to train a model with stability guarantee through variable decorrelation with sample reweighting, taking model misspecification into consideration \cite{shen2020stable,kuang2020stable,lv2023duet,zhang2023map,lv2023ideal}. However, the majority of these approaches are suggested in linear settings. GNNs combine heterogeneous data from node features and graph topological structures, resulting in the existence of intricate and unrecognized non-linear relationships across representations \cite{fan2021generalizing,li2021ood}. Non-linear dependencies on graph data cannot be removed using linear sample reweighting approaches.
Recent studies propose non-linear decorrelation methods for graph tasks \cite{fan2021generalizing,li2021ood}. They attempt to eliminate the dependencies between all the variables of the graph representation through a set of learned sample weights. However, such a demanding aim might result in an excessively small sample size, which hampers the generalization ability of GNNs \cite{martino2017effective,llorente2022optimality,zhang2022tree}. Moreover, these non-linear decorrelation methodologies on graph data suffer from overfitting problems due to the additional hyper-parameters, leading to difficulties in achieving convergence.

We suggest that not all correlations should be eliminated, in contrast to prior techniques \cite{fan2021generalizing,li2021ood}, which aggressively decorrelate all connections across graph representations. Such an aggressive objective may result in an issue with an overly-reduced sample size \cite{martino2017effective,llorente2022optimality}, which hampers the generalization ability of GNNs.
Taking the graph classification task in Figure \ref{fig:fig1} as an example, although varied variables may be used to characterize the ``wheel" motif's graph structure and the features of nodes; they function as an integrated whole, and these relationships remain stable across datasets with varied or unknown distribution shifts.
The significant correlations between the variables in the ``wheel" and the ``star" due to the selection bias seen in the biased training dataset. Such "spurious" relationships, however, cannot be used in OOD datasets. Therefore, to obtain a precise graph model, we just need to eliminate the spurious correlations between two sets of variables (the ``wheel" and the ``star").

In this paper, we propose a framework called L2R-GNN to solve the problem of learning out-of-distribution graph representation. Our framework includes a nonlinear graph decorrelation method that reduces correlations between variables of various clusters. This method is more effective than previous approaches at controlling the overly-reduced sample size and can significantly increase the ability to generalize outside of the distribution. We group the variables of graph representations based on the stability of their correlations and learn a set of weights to remove spurious correlations. By doing so, graph neural networks can focus more on the real relationship between their ground-truth labels and the graph representations. 
We also introduce a stochastic approach based on bi-level optimization for the L2R-GNN framework. This approach allows for the simultaneous learning of the optimal GNN parameters and weights while avoiding the over-fitting problem. Our experimental results show that L2R-GNN outperforms baselines on different graph tasks subjected to distribution shifts. 
Our contributions are as follows: 
1) We propose a novel framework that can learn effective graph representation under complex distribution shifts and achieve better performance simultaneously. 
- We propose a more effective graph decorrelation method than prior approaches at controlling the overly-reduced sample size and increasing the ability to generalize outside of the distribution. 
2) We propose an effective stochastic algorithm based on bi-level optimization for the L2R-GNN framework, which enables simultaneously learning the optimal weights and GNN parameters and avoiding the over-fitting issue. 
3) Our extensive empirical results on several graph benchmarks subjected to distribution shifts show that L2R-GNN greatly outperforms baselines in terms of performance.

\section{Related Works}
\label{section:relatedwork}
 
\textbf{Generalizable Graph Neural Network.}
Most GNNs methods are proposed under the IID hypothesis, which states that the training and testing sets are independently sampled from the same distribution~\cite{kipf2016semi,velivckovic2017graph}. However, in practice, it might be challenging to satisfy this ideal hypothesis. Recent research~\cite{fan2021generalizing,li2021ood} studies how well GNNs generalize outside the training distribution. Several studies concentrate on size generalization ability to make GNNs function effectively on testing graphs whose size distribution is different from that of training graphs. SL-DSGCN \cite{tang2020investigating} reduces the degree-related distribution shifts of GCNs for the OOD node classification task. ImGAGN \cite{qu2021imgagn} produces a set of synthetic minority nodes to balance the class distribution shifts. BA-GNN \cite{chen2022ba} is proposed to learn node representations that are invariant across various distributions for invariant prediction. EERM\cite{wu2022handling} helps GNNs take advantage of invariance principles for prediction on node-level problems. For OOD graph classification task, some works~\cite{fan2021generalizing,li2021ood} improve the generalization capability of GNNs via non-linear decorrelation methods. However, such an aggressive target might result in an issue with an excessively small sample size problem\cite{martino2017effective,llorente2022optimality}, which hampers the generalization ability of GNNs. Moreover, these non-linear decorrelation methods on graph data suffer from over-fitting issues due to the additional hyper-parameters and are hard to converge.

\textbf{The Bi-level Optimization.}
Many works use bi-level optimization~\cite{maclaurin2015gradient,wang2020global,chen2021adaptive,gai2021multi,chen2023mapo} to improve the performance of GNNs. They optimize a higher-level learning subject for lower-level learning. To search the GNN architectures, several studies~\cite{xiao2022decoupled,xiao2023reconsidering,jiang2022role} optimize a bi-level goal using reinforcement learning. Furthermore, \cite{xiao2021learning} presents bi-level programming with variational inference to provide a framework for learning propagation methods. The study~\cite{liu2020towards} attempts to get a parameter initialization that can swiftly adapt to unfamiliar workloads utilizing gradient information from the bi-level optimization. Our study focuses on the capacity of GNNs to generalize in graph-level tasks, and we use bi-level programming to provide a framework for learning graph weights while avoiding over-fitting problems.

\section{Method}
\label{section:method}

\textbf{Problem Formulation.}
Given the training graphs $\mathbf{G}_{train} = \{ G_n, Y_n \}^{N}_{n=1}$,
where $G_n$ is the $n$-th graph and $Y_n$ is the corresponding label.
$\mathbf{G}_{test}$ is the testing graph which is unobserved in the training stage.
The task is to learn a graph neural network  
$GNN({\theta}): \mathcal{G} \rightarrow \mathcal{Z}$ and classifier $\mathcal{R}: \mathcal{Z} \rightarrow \mathcal{Y}$ to predict the label of testing graphs $\mathbf{G}_{test}$, which is under distribution shifts ${\rm P}(\mathbf{G}_{train}) \neq {\rm P}(\mathbf{G}_{test})$.
Denote graph representations $Z = GNN({\theta,G})$, $\mathbf{Z} \subset \mathbb{R}^{N \times d}$. $\mathbf{Z}_{i,j}$ represent the $i$-th row and $j$-th column in Z.

\textbf{Graph Reweighting with RFF}.
Similar to previous works \cite{fan2021generalizing,li2021ood}, we decorrelate graph representations, removing statistical relationships between relevant and irrelevant graph representations. Relevant graph representation is invariant across many unknown testing graphs, while irrelevant representation varies. We remove statistical dependency of all dimensions in representation $\mathbf{Z}$, defined as: 
$\mathbf{Z}_{:,i} \indep \mathbf{Z}_{:,j}, \forall i,j \in [1, d], i \neq j$.
Hypothesis testing statistics evaluate independence between random variables. We use the Hilbert-Schmidt Independence Criterion (HSIC) to supervise feature decorrelation. If the product $k_{\mathbf{Z}_{:,i}}k_{\mathbf{Z}_{:,j}}$ is a characteristic kernel, we have
${\rm HSIC}(\mathbf{Z}_{:,i}, \mathbf{Z}_{:,j}) = 0 \Leftrightarrow \mathbf{Z}_{:,i} \indep \mathbf{Z}_{:,j}.$
However, HSIC is not suitable for training deep models on large datasets due to high computational cost. We use the Frobenius norm as the independent testing statistic for the graph representation space. The partial cross-covariance matrix is:$    \hat{\Sigma}_{\mathbf{Z}_{:,i},\mathbf{Z}_{:,j}} = \frac{1}{N - 1} \sum_{n=1}^N \bigg[ \bigg(\mathbf{u}(\mathbf{Z}_{n,i}) - \frac{1}{N}\sum_{m=1}^N \mathbf{u}(\mathbf{Z}_{m,j})\bigg)^T 
    \cdot \bigg(\mathbf{v}(\mathbf{Z}_{n,j}) - \frac{1}{N}\sum_{m=1}^N \mathbf{v}(\mathbf{Z}_{m,j})\bigg)\bigg]$
where $\mathcal{H}_{\rm RFF}$ represents the space of a random Fourier function. Using $n_u, n_v = 5$ is reliable enough to assess the independence of random variables in real-world situations.

Using the independence criterion, we apply graph reweighting to remove inter-variable dependencies in graph representation, and RFF to assess overall independence.
The learnable graph weight $\mathbf{W} = \{ w_n \}^{N}_{n=1}$ for the $n$-th graph $G_n$ in the training set is $w_n \in \mathbb{R}$. The partial cross-covariance matrix after reweighting is:

\begin{equation}
\begin{aligned}
&\widehat{\Sigma}_{\mathbf{Z}_{:,i}, \mathbf{Z}_{:,j}}^{\mathbf{W}} = \\  
&\quad\frac{1}{N-1} {\sum_{n=1}^{N}} \left[ \left( w_n u(\mathbf{Z}_{ni}) - \frac{1}{N}\sum_{m=1}^{N} w_m u(\mathbf{Z}_{mi})  \right)^\top  \right. \\
&\quad\left. \cdot \left( w_n v(\mathbf{Z}_{nj}) - \frac{1}{N}\sum_{m=1}^{N} w_m v(\mathbf{Z}_{mj})  \right) \right].\label{equation:partial_cross_covariance_weight}
\end{aligned}
\end{equation}

By reducing the squared Frobenius norm of the partial cross-covariance matrix $\lVert \widehat{\Sigma}_{\mathbf{Z}_{:,i}, \mathbf{Z}_{:,j}}^{\mathbf{W}} \rVert_{\rm F}^2$, the optimal graph weight $\mathbf{W}^*$ reduces inter-variable dependencies in graph representation:
\begin{align} 
\label{equation:objective2old} \mathbf{W}^{*}  = {\rm argmin}_\mathbf{W} \sum_{1 \leq i < j \leq d}  \lVert \widehat{\Sigma}_{\mathbf{Z}_{:,i}, \mathbf{Z}_{:,j}}^{\mathbf{W}} \rVert_{\rm F}^2,
\end{align}

Reducing Eq.~\eqref{equation:partial_cross_covariance_weight} directly eliminates correlations between any two variables of graph representation. 

These methods are widely used in recent works \cite{fan2021generalizing,li2021ood}.
However, this aggressive target in  Eq.~\eqref{equation:partial_cross_covariance_weight}  hampers GNNs' generalization capacity due to an excessively decreased sample size problem.

\textbf{Graph Decorrelation}.
We contend that \emph{not all correlations} need to be eliminated, in contrast to prior approaches \cite{fan2021generalizing,li2021ood}, which aggressively decorrelate all dependencies between graph representations.
As an example, consider the graph classification problem given in Figure \ref{fig:fig1}. Although the features of the nodes and the graph structure of the ``wheel'' motif may be represented by several variables, they function as a single unit and exhibit consistent correlations across various unknown testing graphs.
We can see the significant correlations between the variables in the ``wheel" and the ``star" due to the selection bias observed in the biased training dataset. Such "spurious" correlations, however, cannot be used for many unknown testing graphs. In order to get a correct graph model for such a situation, we just need to eliminate the erroneous connection between two sets of data (the ``wheel" and the ``star").

Specifically, we propose a novel nonlinear graph decorrelation method, which is more effective than previous approaches at limiting the overly-reduced sample size and may significantly increase the out-of-distribution generalization ability. The graph decorrelation approach learns a set of weights to reduce correlations between the variables of various clusters rather than any two variables. The variables of graph representation are grouped depending on the stability of their correlations. By eliminating erroneous correlations, the learnt weights enable graph neural networks to focus more on the real relationship between learned discriminative graph representations and their ground-truth labels.

We define the dissimilarity of two variables of graph representation $Z$ as follows in order to express the invariant property of two variables through the variance of their correlation:

\begin{equation}
\label{distance}
\begin{aligned} 
&\operatorname{Dis}\left(Z_{:,i}, Z_{:,j}\right)=\\&\sqrt{\frac{1}{N-1} \sum_{l=1}^{N}\left(\operatorname{Corr}\left(Z_{l,i}, Z_{l,j}\right)-A v e_{-} \operatorname{Corr}\left(\hat{Z}_{:,i}, \hat{Z}_{:,j}\right)\right)^{2}}
\end{aligned}
\end{equation} 
where $A v e_{-} \operatorname{Corr}\left(\hat{Z}_{:,i}, \hat{Z}_{:,j}\right)$ represents the average correlation across whole datasets and $\operatorname{Corr}\left(Z_{l,i}, Z_{l,j}\right)$ represents the pearson correlation of $Z_{l,i}, Z_{l,j}$ in the $l$th graph. 
We load the full dataset in order to obtain $\hat{Z}_{:,i}$ and $\hat{Z}_{:,j}$.  However, because of the high computational cost and enormous storage usage, it is practically impossible to apply in huge datasets. Thus, we propose a scalable method with momentum update.

It makes sense to cluster the variables with lower dissimilarity into the same cluster since they are more likely to retain a stable joint distribution over many graphs.
Therefore, we choose the cluster with the closest mean as determined by the least squared Euclidean distance for each variable in the graph representation $Z$.

We could have $K$ clusters of variables in graph representation $Z$, where $j$th cluster is $S_j$ and cluster center is $\mu_j$.
Thus, we could learn $\mu$ and $S$ by minimizing
\begin{equation} \label{eq:cluCenter}
\mu,S = {\rm argmin }_{\mu,S} \sum^k_j \sum_{{Z}_{:,i} \in S_j} Dis({Z}_{:,i}, \mu_j)
\end{equation}

We could eliminate the correlation between the variables of distinct clusters rather than any two variables by combining the variable clustering of the graph representation. We could reformulate Eq. \ref{equation:objective2old} as:  
\begin{equation}
W^*= {\rm argmin}_{\mathbf{W} }  \sum_{1 \leq i < j \leq d}  \mathbb{I}(i, j)  \lVert \widehat{\Sigma}_{Z_{:,i},Z_{:,j}}^{{\mathbf{W}}} \rVert_{\rm F}^2,
\end{equation}
where indicator variable $ \mathbb{I}(i, j)$ returns 0 if the clusters of $Z_{:,i}$ and $Z_{:,j}$ are different and 1 if the clusters are the same.

To get optimal graph weights $\mathbf{W}$, graph neural network $GNN({\theta})$, and classifier $\mathcal{R}$, we have:
\begin{align}
\label{equation:objective1} \theta^{*},  \mathcal{R}^{*} &= {\rm argmin}_{\theta,  \mathcal{R}} \sum_{n=1}^{N} w_n \ell\left(\mathcal{R} \circ GNN({G,\theta}) , \mathbf{Y}_n\right), \\
\label{equation:objective2} \mathbf{W}^{*} &= {\rm argmin}_\mathbf{W} \sum_{1 \leq i < j \leq d}  \mathbb{I}(i, j)  \lVert \widehat{\Sigma}_{\mathbf{Z}_{:,i}, \mathbf{Z}_{:,j}}^{\mathbf{W}} \rVert_{\rm F}^2,
\end{align}
where $\ell$ is the loss function.  
The statistical dependence between different clusters rather than all variables could be eliminated by jointly optimizing the graph neural network $GNN({\theta})$, classifier $\mathcal{R}$, and graph weights $\mathbf{W}$.

However, such non-linear decorrelation methods on graph data suffer from over-fitting issues due to the additional hyperparameters and are hard to converge.
To overcome such over-fitting problem, we consider a bi-level optimization method for the model framework that allows for the simultaneous learning of the optimal GNN parameters and weights.

\textbf{Bi-level Training Algorithm}. 
We introduce our \model framework to learn effective graph representation. However, as suggested by previous
works \cite{ren2018learning,xiao2021learning}, 
sample reweighting algorithms suffer from over-fitting due to the additional hyperparameters and are hard to converge.
{In this section, we propose the bi-level training algorithm to alleviate the over-fitting problem.}
 
For our proposed \model framework, the introduced decorrelation method for joint learning sample reweights also increases the risk of over-fitting as shown in experiments. 
Inspired by gradient-based
meta-learning (learning to learn), we use bi-level optimization to  solve the over-fitting issue.
Thus, the objective can be formulated as the following bi-level optimization problem:

\begin{linenomath}  
	\begin{align}
	\min _{{W}}  \mathcal{L}_{\text{val}}\left({\theta}^{*}({W}), {W}\right) &=    \sum_{1 \leq i < j \leq d}  \mathbb{I}(i, j)  \lVert {\Sigma}_{ GNN(G_{val},{\theta}^{*}({W}))}^{{\mathbf{W}}} \rVert_{\rm F}^2  \label{Eq:bi-level} \nonumber \\ 
\text { s.t. } {\theta}^{*}({W})&=\arg\min_{{\theta}} \mathcal{L}_{\text {train}}({\theta}, {W}) \nonumber \\ &=    W \ell\left(\mathcal{R} \circ GNN\left(G_{train},\theta \right), \mathbf{Y}_{train}\right), 
	\end{align}
\end{linenomath}

This bi-level update aims to optimize the graph weights based on its validation for avoiding the over-fitting issues, where $\mathcal{L}_{\text {train}}({\theta}, {W})$ and $ \mathcal{L}_{\text{val}}\left({\theta}^{*}({W}), {W}\right)$  are lower-level and higher-level objectives on the training and validation sets, respectively.

Since there is no closed-form expression for $theta$, it is hard to directly optimize the higher-level objective function in Eq.~(\ref{Eq:bi-level}).
We provide an alternating approximation approach to solve such problems.

\textbf{Updating ${\theta}$ in outer loop}. 
Different from previous works \cite{fan2021generalizing,li2021ood}, we do not solve the lower level problem for each outer loop. 
At the $i$-th iteration, we fix ${W}$ and only perform the gradient steps for parameter ${\theta}$ with the learning rate $\eta_{\theta}$ that are listed below:
\begin{linenomath} 
	\begin{align}
	{\theta}^{(i)}={\theta}^{(i-1)}-\eta_{{\theta}} \nabla_{{\theta}} \mathcal{L}_{\operatorname{train}}({\theta}^{(i-1)}, {W}^{(i-1)}), \label{Eq:theta}
	\end{align}
\end{linenomath}

\textbf{Updating ${W}$ in inner loop}. 
We compute the higher-level objective in the inner loop after getting the parameter ${\theta}^{(i)}$, which is an estimate of ${\theta}^{(*)}({W}$)):
\begin{linenomath} 
	\begin{align}
	{W}^{(i)}={W}^{(i-1)}-\eta_{{W}} \nabla_{{W}} \mathcal{L}_{\text{val}}({\theta}^{(i)}, {W}^{(i-1)}). \label{Eq:phi}
	\end{align}
\end{linenomath}

We could have the gradient of $W$, since the $W$ is part of function ${\theta}^{(i)}$ due to Eq. (\ref{Eq:theta}), and the function $\nabla_{W} \mathcal{L}_{\mathrm{val}}(\theta^{(i)}, W^{(i-1)})$ could be represented by:
\begin{align}
&\nabla_{W} \mathcal{L}_{\mathrm{val}}(\theta^{(i)}, W^{(i-1)})=\nabla_{W} \mathcal{L}_{\mathrm{val}}(\bar{\theta}^{(i)}, W^{(i-1)})\label{Eq:gradient} \nonumber \\ 
&-\eta_{\theta}\frac{1}{\epsilon} ( \nabla_{W} \mathcal{L}_{\operatorname{train}}(\theta^{(i-1)}+\epsilon \nabla_{\theta} \mathcal{L}_{\mathrm{val}}(\theta^{(i)}, \bar{W}^{(i-1)}), W^{(i-1)}) \nonumber \\&- \nabla_{W} \mathcal{L}_{\operatorname{train}}(\theta^{(i-1)}, W^{(i-1)})),
\end{align}
where $\bar{\theta}^{(i)}$ and $\bar{W}^{(i-1)}$ denote stopping the gradient. 
Set $\eta_{\theta}$ to 0 in Eq.~(\ref{Eq:gradient}) to obtain a first-order approximation as followed: 
\begin{align}
&\nabla_{W} \mathcal{L}_{\mathrm{val}}(\theta^{(i)}, W^{(i-1)})=\nabla_{W} \mathcal{L}_{\mathrm{val}}(\bar{\theta}^{(i)}, W^{(i-1)})\label{Eq:firstorder}.  
\end{align}

By alternating the update procedures in Eqs.~(\ref{Eq:theta}) and~(\ref{Eq:phi}), we can derive the whole model algorithm from the above gradient derivations. 
Moreover, we investigate the impact of bi-level optimization as well as the first- and second-order approximations in experiments. Results show that the best performance is obtained with a first-order approximation.

\textbf{Momentum Graph Weight Estimator}.
For each graph, a specific weight should be learned as shown in Equation \ref{Eq:bi-level}.
However, simultaneously loading the entire dataset for optimization is impractical due to high computational cost and excessive storage consumption, especially for large datasets.
We employ weight queues with a $K$ dimension to balance optimization performance and weight consistency.
We have graph representation queue $\mathbf{Z}^{(q)}=[\mathbf{Z}^{(q_1)}, \cdots, \mathbf{Z}^{(q_K)}]$ and the corresponding weight queue $\mathbf{W}^{(q)}=[\mathbf{W}^{(q_1)}, \cdots, \mathbf{W}^{(q_K)}]$.
During training, they act as a memory bank from earlier mini-batches.

The graph representations and weights used for optimization are constructed as follows for each mini-batch of input graphs $G_n$:
$\widehat{\mathbf{Z}} =  Concat(\mathbf{Z}^{(q_1)}, \cdots, \mathbf{Z}^{(q_K)}, \mathbf{Z}^{(l)}) , \widehat{\mathbf{W}} = Concat(\mathbf{W}^{(q_1)}, \cdots, \mathbf{W}^{(q_K)}   \mathbf{W}^{(l)})$. 	 
Using graph representations queues, we reformulate Eq.\ref{distance} as:
\begin{equation}
 \label{distanceNew}
\begin{aligned}
&\operatorname{Dis}\left(Z_{:,i}, Z_{:,j}\right)=\\ &\sqrt{\frac{1}{N-1} \sum_{l=1}^{N}\left(\operatorname{Corr}\left(Z^{(l)}_{:,i}, Z^{(l)}_{:,j}\right)-A v e_{-} \operatorname{Corr}\left(\hat{Z}^{(q)}_{:,i}, \hat{Z}^{(q)}_{:,j}\right)\right)^{2}}
\end{aligned}
\end{equation} 
where the pearson correlation of $Z_{:,i}, Z_{:,j}$ in the mini-batch is represented by $\operatorname{Corr}\left(Z^{(l)}_{:,i}, Z^{(l)}_{:,j}\right)$, and their average correlation across all graph representation queues is represented by $A v e_{-} \operatorname{Corr}\left(\hat{Z}^{(q)}_{:,i}, \hat{Z}^{(q)}_{:,j}\right)$.

Our \model reduce the computational cost via weight queues.
If the batch size is $B$, then $\widehat{\mathbf{Z}}$ is a matrix with dimensions of $((k+1) B) \times m_{Z}$, and $\widehat{\mathbf{W}}$ is a vector with dimensions of $(K+1)B $. The computational cost is lowered from $O(N)$ to $O(kB)$ in this manner. 

To dynamically update the representations $\mathbf{Z}^{(q)}$ and weights $\mathbf{W}^{(q)}$ in queues, we use a momentum coefficient $\alpha_i \in [0, 1)$:
$ \mathbf{Z}^{(q_i)*}  =\alpha_{i} \mathbf{Z}^{(q_i)}+\left(1-\alpha_{i}\right) \mathbf{Z}^{(l)}, \mathbf{W}^{(q_i)*} =\alpha_{i} \mathbf{W}^{(q_i)}+\left(1-\alpha_{i}\right) \mathbf{W}^{(l)}$.
We replace all $\mathbf{Z}^{(q_i)}$, $\mathbf{W}^{(q_i)}$ with $\mathbf{Z}^{(q_i)*}$, $\mathbf{W}^{(q_i)*}$ for the next batch.

\begin{table*}[t]
	
	\centering
	\caption{Performance of six Open Graph Benchmark (OGB) graph datasets. 
 }
	\label{table:performance_ogb}
		\resizebox{139mm}{27mm}{
	\begin{tabular}{c|c|c|c|c|c||c }
		\hline
		& \textbf{TOX21}        & \textbf{BACE}         & \textbf{BBBP}         & \textbf{CLINTOX}           & \textbf{HIV}  & \textbf{ESOL}       \\ 
		\hline
		Metric & \multicolumn{5}{c||}{{ROC-AUC ($\uparrow$)}} & \multicolumn{1}{c}{{RMSE ($\downarrow$)}}\\
		\hline

		GIN         & 70.4$\pm$0.9          & 73.8$\pm$3.1          & 67.9$\pm$1.4          & 87.4$\pm$2.8                 & 75.8$\pm$1.2   & 1.14$\pm$0.08 \\
		GCN         & 72.7$\pm$0.6          & {77.6$\pm$1.5}    & 66.8$\pm$1.2          & {88.6$\pm$2.2}           & 76.2$\pm$1.2   & 1.12$\pm$0.04       \\
		SGC   & 71.8$\pm$1.2          & 70.7$\pm$1.4          & 62.7$\pm$1.8          & 76.4 $\pm$2.0            & 67.5$\pm$1.3   & 1.59$\pm$0.06       \\
		
		PNA         & 69.1$\pm$0.8          & 74.9$\pm$1.9          & 64.5$\pm$1.3          & 80.6$\pm$2.3               & {76.2$\pm$1.8}  & {0.98$\pm$0.07}  \\
		JKNet   & 69.8$\pm$1.2          & 77.5$\pm$1.2          & 63.4$\pm$1.1          & 82.4$\pm$2.2            & 73.7$\pm$1.2   & 1.29$\pm$0.09      \\
		SAGPool     & 72.6$\pm$2.5          & 75.8$\pm$1.2          & 68.4$\pm$2.0          & 86.2$\pm$1.3                & 76.4$\pm$1.2    & 1.13$\pm$0.08     \\ 
		TopKPool    & 72.1$\pm$1.5          & 76.5$\pm$2.3          & 67.8$\pm$1.8          & 86.2$\pm$1.2                 & 75.1$\pm$1.2    & 1.10$\pm$0.06       \\ 
		
		FactorGCN   & 56.2 $\pm$2.8          & 68.9$\pm$1.5          & 55.1$\pm$1.6          & 65.7$\pm$2.6            & 57.5$\pm$1.9   & 3.12$\pm$0.17     \\ 
		OOD-GNN     & 76.2$\pm$1.3          & 78.3$\pm$1.8          & 68.4$\pm$2.8          & 89.1$\pm$1.6                & 78.2$\pm$1.2    & 0.94$\pm$0.06      \\ 
		StableGNN     & 74.8$\pm$1.9          & 79.2$\pm$2.4          & 68.1$\pm$2.6          & 88.4$\pm$1.9                & 76.5$\pm$1.8    & 0.96$\pm$0.04      \\ 
		\model    & \textbf{78.6$\pm$1.2} & \textbf{81.9$\pm$1.0} & \textbf{70.6$\pm$1.3} & \textbf{91.9$\pm$1.5}   & \textbf{79.7$\pm$1.0} & \textbf{0.84$\pm$0.07} \\ 
		\hline
	\end{tabular}}
\end{table*}

\begin{table}[t]
	\centering
	\caption{Performance of graph classification accuracy (\%) with graph size distribution shifts, where the training and testing graphs are split by graph sizes. All methods are trained on small graphs and tested on larger graphs. Best results are indicated in bold. 
	}
	\label{table:performance_realworld}
	\resizebox{88mm}{25mm}{
	\begin{tabular}{c|c|c|c}
		\hline
		& \textbf{COLLAB }       & \textbf{PROTEINS }       & \textbf{D\&D }      \\
		\hline 
 
		GIN                 & 56.3$\pm$3.5          & 74.4$\pm$2.5                   & 68.9$\pm$4.1          \\
		GCN                 & 64.0$\pm$2.8          & 74.8$\pm$2.7                 & 71.7$\pm$3.3          \\
		SGC                 & 54.9$\pm$4.1          & 72.6$\pm$2.1                   & 64.2$\pm$3.8          \\
		PNA                 & 58.7$\pm$4.3          & 71.9$\pm$2.9              & 70.5$\pm$2.4          \\
		
		JKNet                 & 57.9$\pm$3.8          & 74.2$\pm$2.6                   & 69.5$\pm$3.6          \\

		SAGPool             & {66.8$\pm$1.3}    & {75.9$\pm$0.6}       & {77.1$\pm$1.6}     \\
		TopKPool            & 54.7$\pm$1.4          & 65.7$\pm$2.8          & 68.2$\pm$3.2          \\
		FactorGCN           & 52.3$\pm$1.8          & 62.4$\pm$4.3                 & 55.2$\pm$2.4          \\
		OOD-GNN             & {66.9$\pm$1.5}    & {77.1$\pm$1.1}    & {79.1$\pm$1.3}     \\
		StableGNN             & {67.3$\pm$1.4}    & {76.5$\pm$0.9}     & {78.7$\pm$1.6}     \\
		
		\model            & \textbf{68.2$\pm$1.4} & \textbf{78.9$\pm$0.7}  & \textbf{80.8$\pm$1.2} \\
		\hline
	\end{tabular}}
\end{table}

\begin{figure}[t]
	\centering
	\begin{subfigure}[t]{0.48\linewidth}
		\centering
		\includegraphics[width=1.09\textwidth]{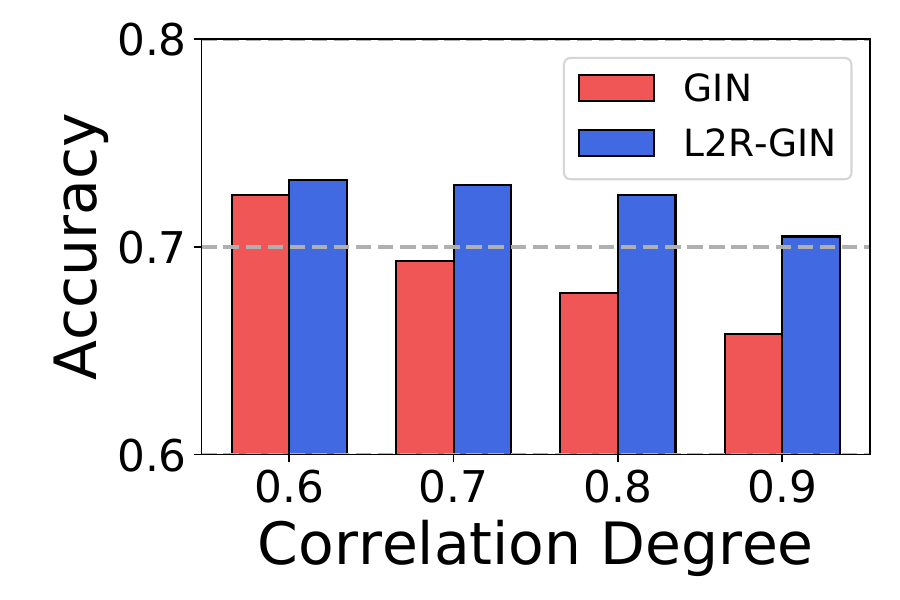}
	\end{subfigure}
	~ 
	\begin{subfigure}[t]{0.48\linewidth}
		\centering
		\includegraphics[width=1.09\textwidth]{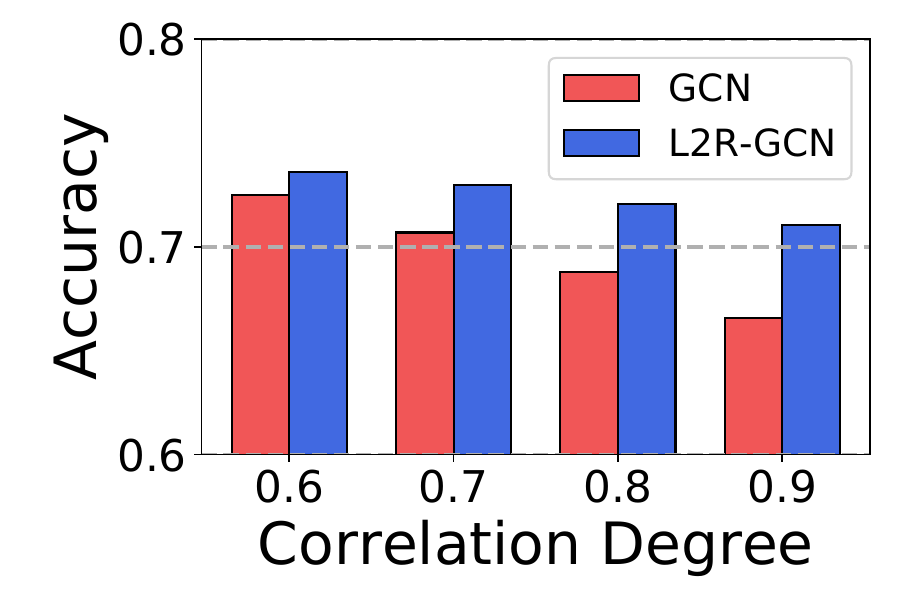}
	\end{subfigure}
	\caption{Results of GCN and GIN backbones under different correlation degree settings.
		Comparing with GCN and GIN method, our L2R-GNN methods (by applying our L2R-GNN framework on GCN and GIN backbone) improves the
		accuracy of graph classification across different spurious correlation degree settings.}
	\label{figure:compareBack}
\end{figure}

\begin{figure}[t]
	\centering
	\begin{subfigure}[t]{0.45\linewidth}
		\centering
		\includegraphics[width=1.09\textwidth]{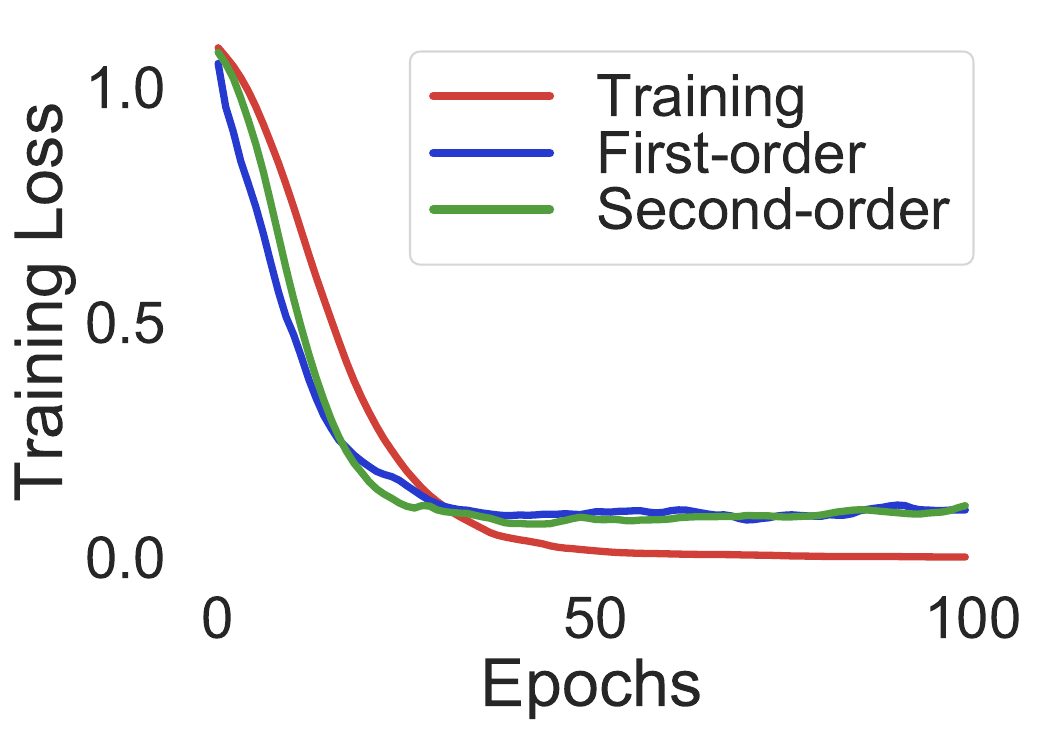}
	\end{subfigure}
	~ 
	\begin{subfigure}[t]{0.45\linewidth}
		\centering
		\includegraphics[width=1.09\textwidth]{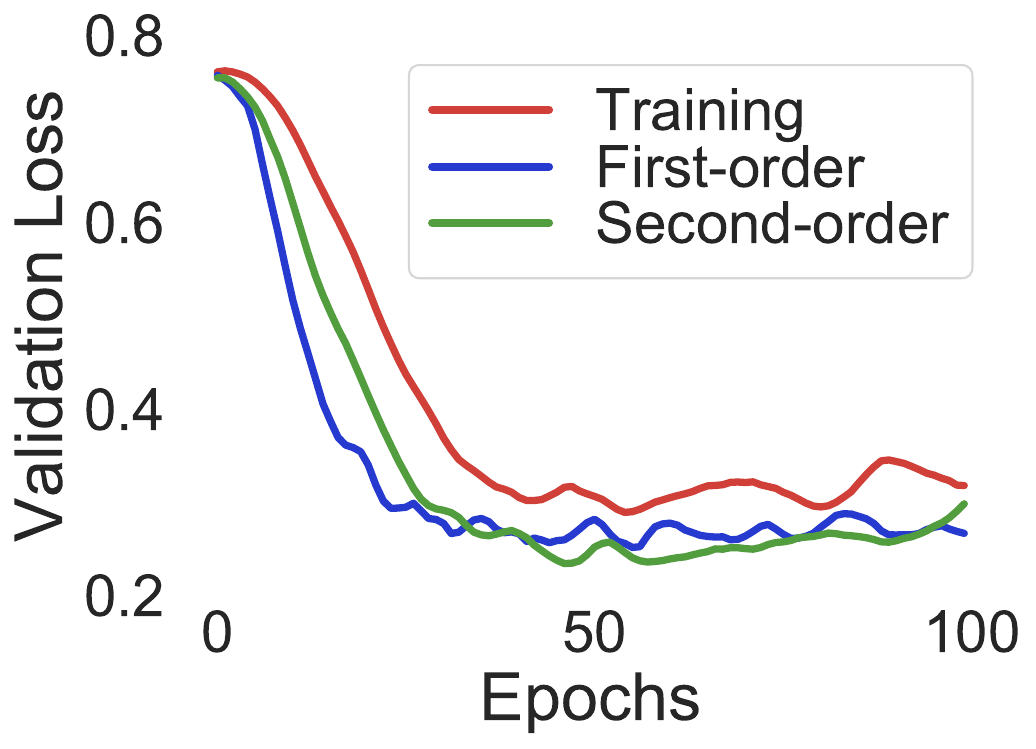}
	\end{subfigure}
\vspace{-1em}
	\caption{The training and validation loss curves on D\&D.}
	\label{figure:losscurve}
\end{figure}

\begin{figure*}[t]
	\centering
	\includegraphics[width=0.99\textwidth]{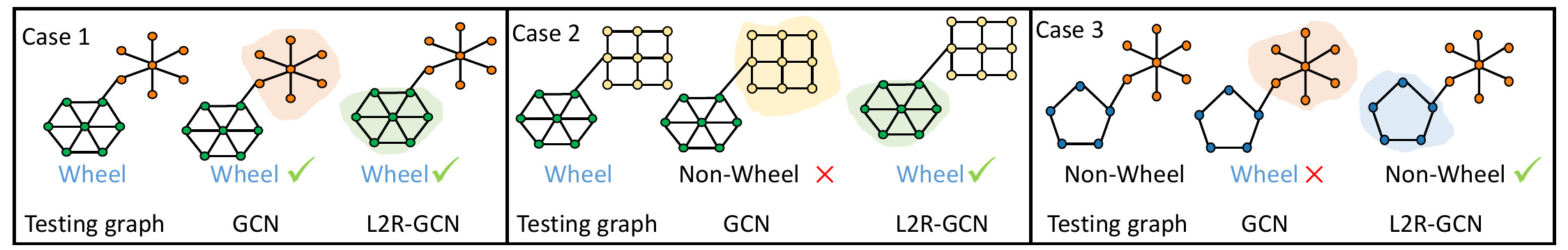}
	\vspace{-1em}
	\caption{ Case study of GCN and \model  in biased synthetic dataset. Shadow areas are the important subgraph calculated by the GNNExplainer.}
	\label{fig:syn_explainer}
\end{figure*}

\begin{figure}[t]
	\centering
	\begin{subfigure}[t]{0.45\linewidth}
		\centering
		\includegraphics[width=1.09\textwidth]{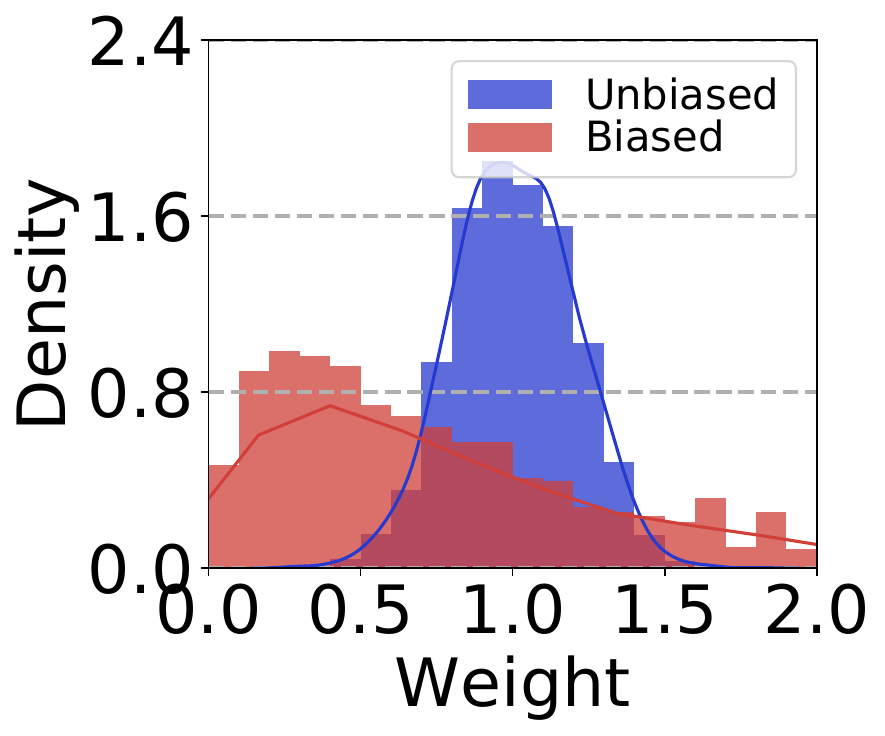}
		\caption{Synthetic dataset.}
	\end{subfigure}
	~ 
	\begin{subfigure}[t]{0.45\linewidth}
		\centering
		\includegraphics[width=1.09\textwidth]{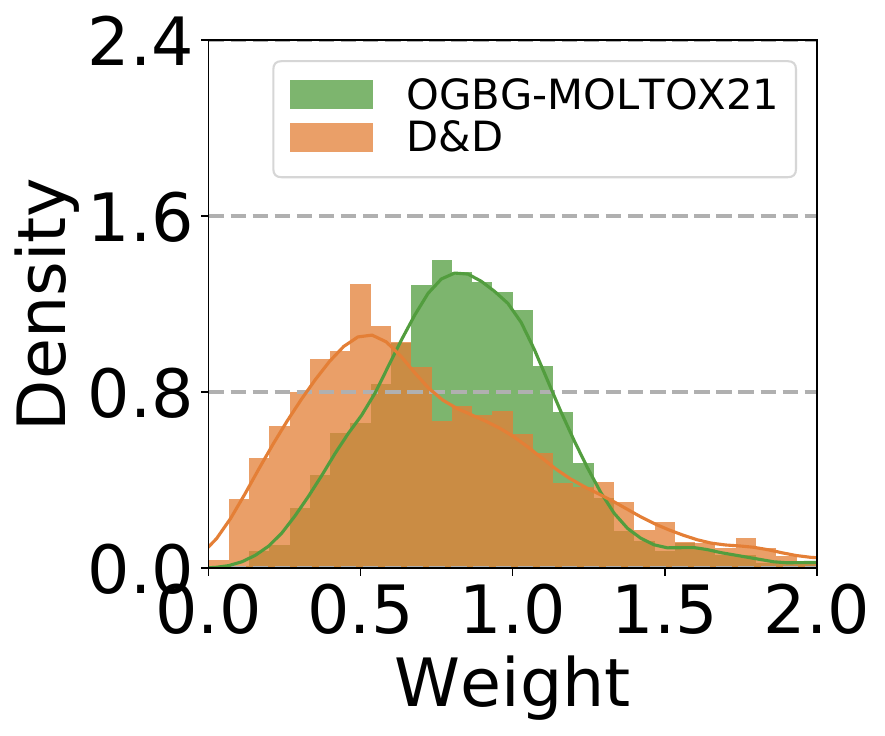}
		\caption{OGBG and D\&D datasets.}
	\end{subfigure}
\vspace{-1em}
	\caption{The distribution of the learned graph weights on (a) unbiased and biased synthetic dataset and (b) two real-world datasets.}
	\label{figure:weight_distribution}
\end{figure}

\begin{figure}[t]
	\centering
	\begin{subfigure}[t]{0.45\linewidth}
		\centering
		\includegraphics[width=1.09\textwidth]{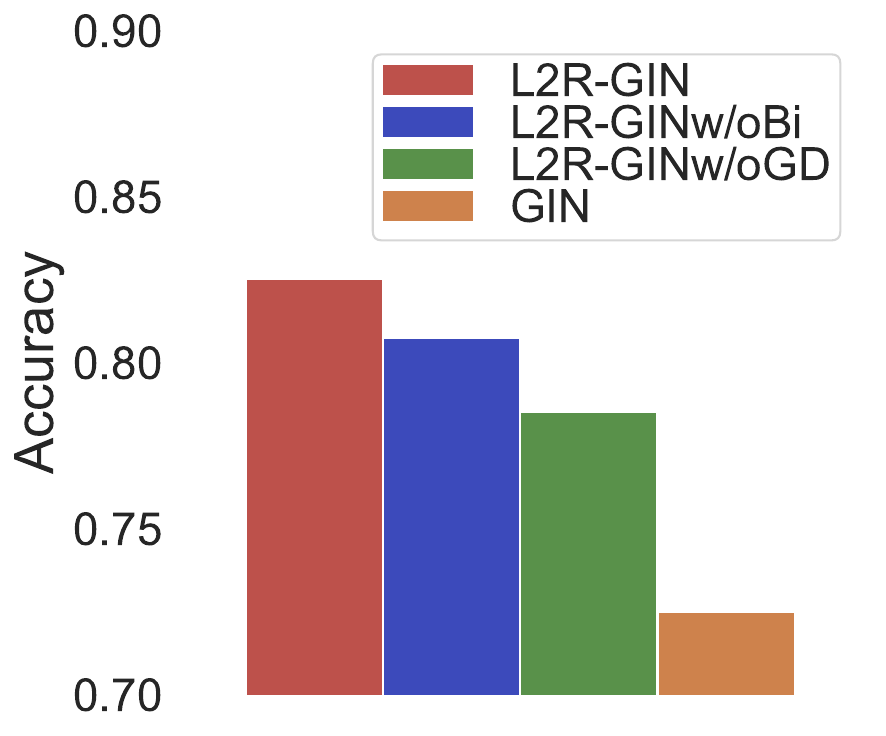} 
	\end{subfigure}
	~ 
	\begin{subfigure}[t]{0.45\linewidth}
		\centering
		\includegraphics[width=1.09\textwidth]{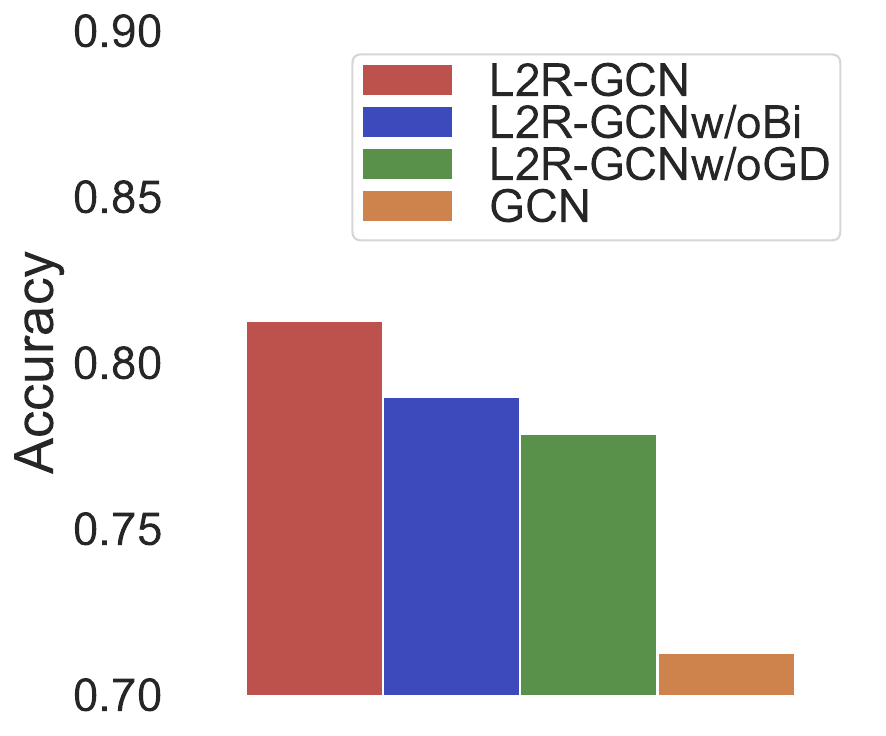} 
	\end{subfigure}
\vspace{-1em}
	\caption{Ablation study of our \model with (a) GIN and (b) GCN backbones.}
	\label{figure:abl}
\end{figure}

\section{Experiments}
\label{section:experiments}

In this section, we describe the experimental setup used to evaluate the effectiveness of our proposed method.
Experimental results demonstrate the effectiveness of our framework in comparison with different GNN backbones and datasets.  
We specifically aim to answer the following questions: 
\textbf{(RQ 1)} How effective is the proposed \model framework for the graph classification task?
	 \textbf{(RQ 2)}  Could the proposed \model  alleviate  different distribution shifts?
	 \textbf{(RQ 3)}  Could the proposed Bi-Level Training Algorithm alleviate the over-fitting issue? 
	 \textbf{(RQ 4)} Does the proposed reweighting mechanism work as designed and give some useful insights?
	 \textbf{(RQ 5)} What are the effects of our proposed different components?

\subsubsection{Baselines}

We compare our \model with several representative state-of-the-art methods: GCN~\cite{kipf2016semi}, GIN~\cite{xu2018powerful}, SGC~\cite{pmlr-v97-wu19e}, JKNet~\cite{xu2018representation}, FactorGCN~\cite{yang2020factorizable}, PNA~\cite{corso2020principal}, TopKPool~\cite{gao2019graph}, SAGPool~\cite{lee2019self}, OOD-GNN~\cite{li2021ood} and StableGNN~\cite{fan2021generalizing}.

\subsubsection{Datasets}
\label{section:datasets}

We evaluate our method and baselines on synthetic and real-world datasets for complex and realistic graph distribution shifts: 
\textbf{Synthetic Datasets.} 
To validate \model effectiveness with various distribution shifts, we generate synthetic datasets, allowing biased degree creation.
Following GNN explanation works~\cite{ying2019gnnexplainer,lin2021generative},
we focus on graph classification tasks with distribution shifts from training to testing datasets. We create a base subgraph for each graph, where each positive graph has a ``wheel"-structured network motif and each negative graph has a motif chosen from four candidates: ``star", "circle", "grid", and "diamond". The ``wheel" motif is the causal structure determining the label.

\textbf{Real-world Datasets.} 
(1) {Molecule and social datasets}.    
Similar to previous works \cite{knyazev2019understanding}, we consider three graph classification benchmarks: COLLAB, PROTEINS, and D\&D. These datasets are split based on graph size.  
Methods are trained on smaller graphs and tested on unseen larger graphs. Specifically, {COLLAB} is a social dataset with 3 public datasets: High Energy Physics, Condensed Matter Physics, and Astro Physics. We train on graphs with nodes from $32$ to $35$ and test on graphs with nodes from $32$ to $492$. {PROTEINS} is a protein dataset. We train on graphs with nodes from $4$ to $25$ and test on graphs with nodes from $6$ to $620$. {D\&D} is also a protein dataset. 
We train on graphs with nodes from $30$ to $300$ and test on graphs with nodes from $30$ to $5,748$.  
(2){Open Graph Benchmark (OGB)}~\cite{hu2020open}. 
We consider {OGBG-MOL$*$} TOX21, BACE, BBBP, CLINTOX, HIV, and ESOL as six graph property prediction datasets from OGB with distribution shifts. Predicting target molecule properties is the graph classification task. 
We use scaffold splitting technique to separate graphs based on two-dimensional structural frameworks. This technique divides structurally diverse molecules into subsets, creating a more realistic and hard out-of-distribution generalization situation.

\textbf{RQ1. Performance Comparison}.
Results on 6 OGB datasets are in Table~\ref{table:performance_ogb}.
Datasets use scaffold splitting~\cite{wu2018moleculenet}, dividing molecules by 2D structural frameworks, creating distribution shifts between train and test graphs.
We could find that \model outperforms other GNN models in all cases, effectively alleviating distribution shifts.
 \model surpasses StableGNN and OOD-GNN, showing effectiveness of graph decorrelation method and bi-level optimization.
\model performs well on various tasks and dataset scales, indicating generality.
 \model excels in out-of-distribution generalization, esp. for large-scale real-world graphs. 
Size generalization problem considered on real-world molecule and social datasets (COLLAB, PROTEINS, D\&D), with train and test graphs split by size. Results in Table~\ref{table:performance_realworld}. \model outperforms baselines, demonstrating best out-of-distribution generalization under size distribution shifts.

\textbf{RQ2. Different Distribution Shifts}.
To inject spurious correlation, $\mu * 100\%$ of ``wheel" graphs have ``star" motif added, and the remaining graphs have a non-causal motif chosen from 4 candidate motifs. For all nodes, node features are taken from the same uniform distribution. To create four spurious correlations for the training set, we set $\mu$ as \{0.6, 0.7, 0.8, 0.9\}. To answer RQ2, we conduct experiments on synthetic datasets with different distribution shifts. The results of GCN and GIN backbones under different correlation degree settings are shown in Figure \ref{figure:compareBack}.  As spurious correlation degree increases, both GIN and GCN experience significant performance decline, suggesting spurious correlation greatly impacts GNNs' generalization performance, and larger distribution changes result in greater performance decline.
Our L2R-GNN methods, implementing our framework on GCN and GIN backbones, increase graph categorization accuracy across various spurious correlation degree settings compared to GCN and GIN methods. The ``wheel" motif, as described earlier, represents the label; thus, using this causal subgraph is the only way to improve performance. This demonstrates how our models can significantly reduce the impact of erroneous subgraph correlation.
 In all biased scenarios, our \model outperforms backbones like GCN and GIN, proving it is a universal framework capable of fitting different GNN architectures.

\textbf{RQ3. Bi-Level Optimization}. To answer \text{RQ3}, we conduct experiments to analyze the model loss during training. We use \textit{training} as a baseline, where we optimize $W$ simultaneously with $\theta$ on training data without validation. We compare training with \textit{first-order} and \textit{second-order} approximates. 
Figures~\ref{figure:losscurve} shows the learning curves of training loss and validation loss on the D\&D dataset of \model.
We can observe that the \textit{training} gets stuck in the over-fitting issue attaining low training loss but high validation loss. For first-order and second-order, the difference between training and validation losses is significantly lower. It proves that the first-order approximation is adequate to prevent over-fitting and the bi-level optimization may greatly increase generalization capabilities.

\textbf{RQ4. Reweighting Mechanism}.
We study the reweighting mechanism's contribution to robust graph representation learning via experiments on synthetic and real-world datasets. In synthetic datasets, $\mu * 100\%$ positive graphs have a "star" motif added, and the remaining positive and negative graphs have a non-causal motif chosen from 4 candidates. We collect graph weights in unbiased ($\mu = 0.25$) and biased ($\mu = 0.8$) datasets, as shown in Figure \ref{figure:weight_distribution}(a). Median weight in biased data is lower than in unbiased data, indicating \model can identify noise graphs with spurious correlation.
 Weight variance in biased data is higher than in unbiased data, showing \model's reliable detection of noise graphs with spurious correlation.
On real-world datasets D\&D and OGBG-MOLTOX21, Figure \ref{figure:weight_distribution}(b) displays learned graph weight distribution, demonstrating non-trivial weights and varying distribution across datasets.

\textbf{Case study.} Using GNNExplainer~\cite{ying2019gnnexplainer}, we visualize important subgraphs for GNN's prediction as shadow areas and compare GCN with \model in Figure \ref{fig:syn_explainer}. Three cases demonstrate L2R-GNN effectiveness:
\emph{Case 1.} GNNExplainer~\cite{ying2019gnnexplainer} shows GCN assigns higher weights to "star" motif, while \model focuses on "wheel" motif. GCN's accurate but unstable prediction may rely on spurious correlation, which is undesirable.
    \emph{Case 2.} GCN ignores "wheel" motif due to spurious correlation, leading to incorrect prediction. \model focuses on "wheel" motif, determining the true label.
\emph{Case 3.} Spurious correlation causes GCN to focus on "star" motif and make incorrect predictions. \model's decorrelation of subgraphs attributes more to prediction with "circle" motif. 

\textbf{RQ5. Component Effects}
We conduct an ablation study and hyper-parameter sensitivity analysis to understand component effects on performance. We compare \model with:
 \textbf{L2R-GNNw/oBi}: L2R-GNN without bi-level optimization, optimizing $W$ and $\theta$ simultaneously on training data without validation.
 \textbf{L2R-GNNw/oGD}: L2R-GNN without graph decorrelation module. 
Results in Figure \ref{figure:abl} show \model achieves the best performance, indicating each component contributes to effectiveness and robustness. Both components contribute to performance gain, complementing each other.

\section{Conclusions}
\label{section:conclusions}

GNNs achieve state-of-the-art performance in tasks like molecular graph prediction, scene graph classification, and social network classification.
We propose \framework for OOD generalization of GNNs.
Our novel nonlinear graph decorrelation method improves OOD generalization and outperforms previous methods in preventing over-reduced sample size.  
We propose a bi-level optimization-based stochastic algorithm for the \model framework, enabling simultaneous learning of optimal example weights and GNN parameters, and avoiding over-fitting.
Empirical results on synthetic and real-world datasets demonstrate \model's effectiveness.

\section{Acknowledge}
This work was supported in part by Young Elite Scientists Sponsorship Program by CAST (2021QNRC001), National Natural Science Foundation of China (No. 62376243, U20A20387), the StarryNight Science Fund of Zhejiang University Shanghai Institute for Advanced Study (SN-ZJU-SIAS-0010), Project by Shanghai AI Laboratory (P22KS00111) and Program of Zhejiang Province Science and Technology (2022C01044).

\bibliography{kdd_conference}

\end{document}